\documentclass{article}
\usepackage[authoryear,round,numbers]{natbib}
\usepackage{PRIMEarxiv}
\usepackage[hidelinks]{hyperref}
\usepackage[utf8]{inputenc} % allow utf-8 input
\usepackage[T1]{fontenc}    % use 8-bit T1 fonts
\usepackage{hyperref}       % hyperlinks
\usepackage{url}            % simple URL typesetting
\usepackage{booktabs}       % professional-quality tables
\usepackage{amsfonts}       % blackboard math symbols
\usepackage{nicefrac}       % compact symbols for 1/2, etc.
\usepackage{microtype}      % microtypography
\usepackage{lipsum}
\usepackage{fancyhdr}       % header
\usepackage{graphicx}       % graphics
\graphicspath{{media/}}     % organize your images and other figures under media/ folder
\usepackage{tabularray}
\title{Benchmarking with MIMIC-IV, an irregular, sparse clinical time series dataset
}

\author{
  Hung Bui \\
  HCLTech \\
  Hanoi, Vietnam\\
  \texttt{hung.bui@hcl.com} \\
  \And
  Harikrishna Warrier \\
  HCLTech \\
  Bangalore, India\\
  \texttt{harikrishna.w@hcl.com} \\
   \And
  Yogesh Gupta \\
  HCLTech \\
  Noida, India\\
  \texttt{yogeshg@hcl.com} \\
}

\begin{document}
\maketitle

\begin{abstract}
%\lipsum[1]
Electronic health record (EHR) is more and more popular, and it comes with applying machine learning solutions to resolve various problems in the domain. This growing research area also raises the need for EHRs accessibility. Medical Information Mart for Intensive Care (MIMIC) dataset is a popular, public, and free EHR dataset in a raw format that has been used in numerous studies. However, despite of its popularity, it is lacking benchmarking work, especially with recent state-of-the-art works in the field of deep learning with time-series tabular data. The aim of this work is to fill this lack by providing a benchmark for latest version of MIMIC dataset, MIMIC-IV. We also give a detailed literature survey about studies that has been already done for MIIMIC-III.
\end{abstract}

% keywords can be removed
\keywords{Electronic Health Record \and MIMIC-IV \and Time Series \and Mortality \and Length of Stay}

\section{Introduction}
%\lipsum[2]
%\lipsum[3]
Irregularly sampled time series data occur in multiple scientific and industrial domains including finance, climate science and healthcare. In healthcare, electronic health records (EHR) have been widely adopted with the hope they would save time and improve the quality of patient care. The role of Artificial Intelligence (AI) in EHR is rapidly transforming the healthcare landscape, offering new opportunities to improve patient care, enhance decision-making, and optimize healthcare operations \citet{shukla2022}. Time-series data is routinely collected in various healthcare settings where different measurements are recorded for patients throughout their course of stay. Predicting clinical outcomes like mortality, decompensation, length of stay, and disease risk from such complex multivariate time-series data can facilitate both effective management of critical care units and automatic personalized treatment recommendations for patients \citet{tipir2022}. However, modeling time series data subject to irregular sampling poses a significant challenge to machine learning models that assume fully observed, fixed-size feature representations \citet{shukla2021multitime}.
MIMIC is mentioned as a pioneer and prominent large-scale openly available EHR dataset that addresses the privacy issues of EHR data by carefully de-identifying patient information. MIMIC-III is a freely available database with 58,976 admissions from 48,520 patients who stayed in critical care units of the Beth Israel Deaconess Medical Center between 2001 and 2012. The database records information such as demographics, vital sign measurements, laboratory test results, medications, imaging reports, etc. MIMIC-III has played an important role in inspiring large amounts of scientific works in clinical informatics, artificial intelligence, and epidemiology.
The latest version of the dataset, MIMIC-IV is an update to MIMIC-III, which incorporates contemporary data and improves on numerous aspects of MIMIC-III. It adopts a modular approach to data organization, highlighting data provenance and facilitating both individual and combined use of disparate data sources. MIMIC-IV is aimed to continue the success of MIMIC-III and provide a broad set of healthcare applications. The MIMIC-IV comprises EHRs from the patients admitted at Beth Israel Deaconess Medical Center in Boston, MA, USA, from the years 2008 to 2019. It contains data of 383,220 patients admitted to an intensive care unit (ICU) or the emergency, which is considerably larger than previous MIMIC-III.
In this paper, we present a study that verifies different algorithms for predicting irregular, sparse clinical data with MIMIC-IV, and provide a detailed benchmark for it. The remainder of this paper is arranged as follows: 
\begin{itemize}
\item Related work: Brief review some of deep learning models that have been trained with MIMIC dataset and then describe some similar work that present benchmark for MIMIC-III, and different data pipeline for MIMIC III and IV dataset.
\item Experiment: Describe training pipeline, models and their hyperparameters that will be included in the experiment. Then, give conclusion about the training result. 
\end{itemize}
\section{Related work}
 The EHR data are known to be messy, often consisting of high-dimensional, irregularly sampled time series with multiple data types and missing values, which requires large effort to build systematic, producible extracting and preprocessing pipeline. Many works have tried to reduce the time and effort spent on labor-intensive data extracting/preprocessing by providing a standard pipeline for extracting and processing the MIMIC data. MIMIC-EXTRACT, \citet{10.1145/3368555.3384469} is a popular pipeline for MIMIC-III that uses a larger set of variables and includes ventilation, vasopressors, and fluid bolus therapies, expanding the tractability for downstream task. \citet{10.1093/jamia/ocaa139} introduce FIDDLE (Flexible Data-Driven Pipeline), that developed generalizable tools for EHR feature extraction, that can be applied for both MIMIC and eICU dataset \citet{Pollard2018TheEC}. However, both MIMIC-EXTRACT and FIDDLE are applicable for MIMIC-III dataset only.  \citet{jarrett2021clairvoyance}, proposes a unified, end-to-end, autoML-friendly pipeline that serves as a software toolkit, empirical standard, and interface for optimization. However, Clairvoyance does not assist in cohort extraction steps but provides different preprocessing and imputation techniques for time-series medical data, and it also works with MIMIC-III only. A recently proposed pipeline for MIMICIV is COP-E-CAT \citet{10.1145/3459930.3469536} that can configure preprocessing steps, providing a user some flexibility to define cohorts as needed for analysis tasks. The limitation of this work is that it does not support non-ICU data, and it leaves out readmission, and LOS prediction tasks. One of the recent works is  \citet{gupta2022extensive}, that provides a pipeline for MIMIC-IV. This work has different modules, forming end-to-end pipeline: data extraction, data preprocessing, predictive modelling, and evaluation. It also provides add-on features such as model calibration and fairness evaluation. However, the number of supported predicting models in this pipeline is limited. Another recent work is presented by \citet{LIAO2023104356} which supports both MIMIC-IV and ICU datasets, the two most popular public clinical datasets. However, this work only provides predictions for in-hospital mortality, acute respiratory failure and shock which excludes length of stay predictions. 

The success of deep learning in image and text domains realized by convolutional and recurrent networks, and transformer models have inspired the application of these architectures to develop better prediction models for irregularly time-series data as well. Many researchers have adopted these models for clinical prediction using MIMIC dataset.  A straight-forward approach to deal with irregular time intervals and missingness is to aggregate measurements into discrete time intervals and add missingness indicators, respectively, before feeding them to a classifier. Existing sequence models address this issue by including a learnable imputation or interpolation component. \citet{shukla2021multitime} proposed a VAE-based (Variational Auto Encoder) architecture with a Multi-Time Attention Networks (mTAN) module. Their primary innovations are the inclusion of a learned continuous-time embedding mechanism coupled with a multi-head temporal cross attention encoder and decoder module (the mTAND module) to provide the interface to produce a fixed-length representation of a time series containing a variable number of observations. The limitation of mTAN is that it uses a homoscedastic output distribution that assumes constant uncertainty and the mTAN model’s cross attention operation normalizes away information about input sparsity. It becomes problematic in cases where there is variable input density through time resulting in the need for encoding, propagating, and reflecting that uncertainty in the output distribution. The Heteroscedastic Temporal Variational Autoencoder (HeTVAE) \citet{shukla2022} addresses mTAN’s issues by introducing an uncertainty-aware multi-time attention network (UnTAN). The UnTAN layer utilizes an attention mechanism to generate a distributed latent representation of irregularly sampled time series at a set of reference time points, which can more directly encode information about input uncertainty. However, both works only consider a very small number of time series data features in MIMIC dataset for training, and also exclude its demographics and static features.
On the other hand, Self-supervised Transformer for Time-Series (STraTS) \citet{tipir2022} model, does not use imputation/interpolation scheme, overcomes these pitfalls by treating time-series as a set of observation triplets, in which a novel Continuous Value Embedding technique is utilized to encode continuous time and variable values without the need for discretization. The model also includes a transformer component with multi-head attention layers, helping it to learn contextual triplet embeddings. Another model is worth to mention is Temporal Convolutional Networks (TCNs) \citet{8099596}, that combines best practices such as dilations and residual connections with the causal convolutions, which can take a sequence of any length and map it to an output sequence of the same length, just as with an RNN. \citet{tcn_mimic3} claimed TCN models substantially outperform generic recurrent architectures such as Long Short-Term Memory (LSTMs) and Gate Recurrent Unit (GRUs) on MIMIC-III dataset. Other methods modify traditional recurrent current network architecture for clinical time-series to handle missing values and/or irregular time intervals. For example, \citet{10.1145/3097983.3097997} developed a time-aware long short-term memory (T-LSTM) which is a modification of the LSTM cell to adjust the hidden state according to the irregular time gaps. ODE-RNN (Ordinary Differential Equation – Recurrent Neural Network) \citet{NEURIPS2019_42a6845a} uses ODEs to model the continuous-time dynamics of the hidden state while also updating the hidden state at each observed time point by utilizing a standard GRU layer. The GRU-D model \citet{rnn_scir8} is a modification of the GRU cell which decays inputs (to global means) and hidden states through unobserved time intervals. DATA-GRU \citet{Tan_Ye_Yang_Liu_Ma_Yip_Wong_Yuen_2020}, in addition to decaying the GRU hidden state according to elapsed time, also employs a dual attention mechanism to preprocess the training data before feeding them to a GRU cell.

Regarding benchmarks for MIMIC dataset, there are couples of works that benchmarks on MIMIC-III dataset. \citet{sanjay2018} provided three clinical prediction tasks including mortality prediction (in-hospital prediction, short-term mortality prediction, long-term mortality prediction), ICD-9 code group prediction (ICD stands for International Statistical Classification of Diseases and Related Health Problems), and length of stay prediction. In this work, they include different prediction algorithms for benchmarking tasks such as SOFA (Sequential Organ Failure Assessment) and SAPS II (Simplified Acute Physiology Score). Other supervised learning algorithms are also provided, including logistic regression, random forest, and deep learning such as Feedforward Neural Network and Recurrent Neural Network. They also provide an ensemble of FFN and GRU deep learning models to learn shared representations from multiple modalities.  A more recent work is from  \citet{hrayr2019} in which they focused on Logistic Regression and LSTM-based algorithms with 4 clinical prediction tasks: in-hospital mortality, physiologic decompensation, length of stay (LOS), and phenotype. They also formulate a heterogeneous multitask learning problem that involves jointly learning all four prediction tasks simultaneously. Beside the numbers of supported prediction tasks between \citet{sanjay2018} and \citet{hrayr2019}, there is one critical different is that in \citet{sanjay2018} the data input is either the first 24 or 48 hours for all prediction tasks, while in \citet{hrayr2019}, they do LOS and decompensation prediction at each hour of the stay, and do phenotyping based on the data of the entire stay. In addition, in \citet{hrayr2019}, they frame the LOS prediction as a classification problem and use Cohen’s kappa score, whereas in \citet{sanjay2018}, they frame it as a regression problem and use mean squared error as its metric. The limitation of these two works is that the scoring methods (SAPS II, SOFA) was not efficient and failed predict the mortality, and these machine learning and deep learning algorithms are also outdated and does not compete with current state of the art models in term of time series prediction for sparse and irregular clinical data. Another thing is that these works do benchmark for MIMIC-III dataset only. All mentioned works were implemented with different data extraction, data preprocessing, evaluation metrics. As a result, it is needed to have standardized benchmarking with same data pipeline for verifying those algorithms.
In our work, we present detailed benchmarking results of deep learning models on MIMIC-IV dataset for two clinical prediction tasks including mortality prediction and length of stay prediction. The in-ICU mortality prediction is predicting the risk of in-ICU death in patient, whereas the patient ICU length of stay can be defined as the number of days that an in-patient will remain in ICU during a single admission by subtracting day of admission from day of discharge. We formulate the length of stay problem as a binary classification whether the patient’s length of stay is smaller than 3 days or not. We do not consider readmission prediction since it is required to process the notes which plays an important role for model to learn \citet{hsu-etal-2020-characterizing}. 

\section{Experiments}
\subsection{MIMIC-IV dataset}
MIMIC-IV is the result of a collaboration between Beth Israel Deaconess Medical Center (BIDMC) and Massachusetts Institute of Technology (MIT), which was first released with version 0.3 in August 2020. This work is using MIMIC-IV 2.2, which was released in January 2023. The data were sourced from two in-hospital database systems: a custom hospital wide HER and an ICU specific clinical information system, which are included in two separate modules hosp and icu respectively, to highlight their provenance. Previously, in MIMIC-III, the whole data was given as one large set, with no obvious differentiation between them. This modular organization allows linking of the database to external departments and distinct modalities of data. All modules can be linked by identifiers such as subject\_id, hadm\_id, and de-identified data and time. Those two modules have been included in MIMIC-III. MIMIC-IV has also improved their data coverage by renaming icustay\_id with stay\_id, which is used to index the stays across different areas of the hospital, i.e a stay in the emergency department, ICU, and operating room will all be distinct and referred to by the same identifier. As a result, if a table data were collected from the ICU clinical information system, it will only provide data for patients while they are in an ICU. In addition, MIMIC-IV is collected from 2008 to 2019, rather than from 2001 to 2012 like MIMIC-III, which is updated with more recent biomarkers. MIMIC-IV also has table-wise improvements over MIMIC-III, that can be seen in \citet{mimic422}. MIMIC-IV encompasses both ICD-9 and ICD-10 codes, unlike MIMIC-III, which exclusively contains ICD-9 codes. The dataset includes 209,359 hospital admissions with ICD-9 codes and 122,317 hospital admissions with ICD-10 codes, compared to MIMIC-III's 52,726 documents \citet{nguyen2023mimicivicd}.
\subsection{Data preparation}
\begin{figure}
    \centering
    \includegraphics[width=1\linewidth]{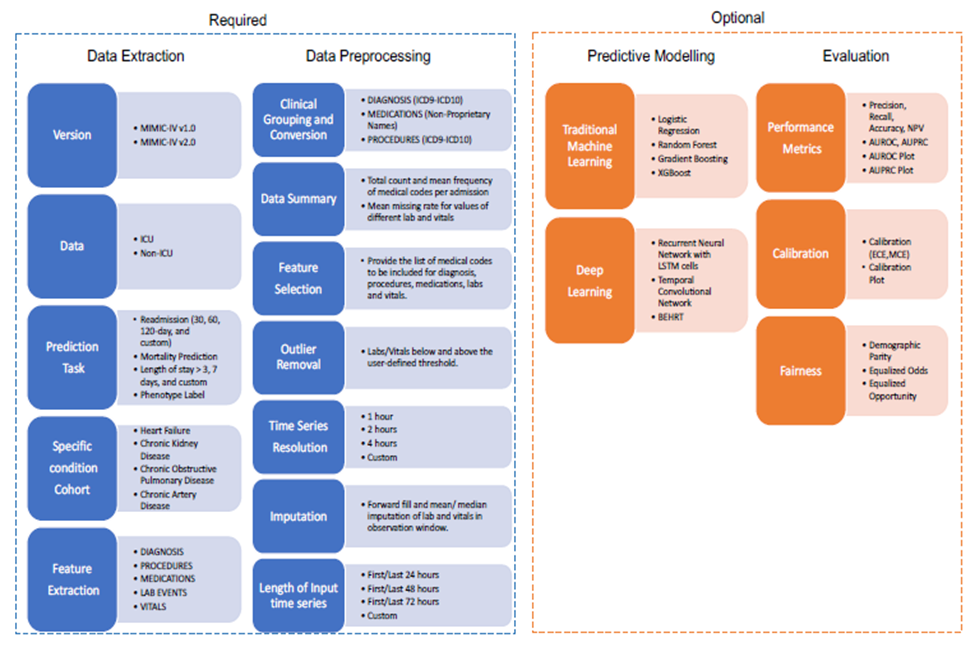}
    \caption{Pipeline Overview from \citet{gupta2022extensive}}
    \label{fig:fig_pipe}
\end{figure}
In this work, the MIMIC-IV dataset is extracted and preprocessed by using the work from \citet{gupta2022extensive}. They provide the complete pipeline from extracting data, preprocessing, training, and evaluation which is described in Figure \ref{fig:fig_pipe}. This pipeline supports MIMIC-IV 2.0, however it is also able to be used with the latest, MIMIC-IV 2.2. Regarding evaluation, we utilize area under the ROC curve (OC-AUC score) and area under precision-recall curve (PR-AUC score). One of the reasons for selecting this pipeline for our work is its extensibility which helps us to customize the pipeline based on our preference. Another reason is that they provide rich features for extracting and preprocessing data, especially for time series features. 
We used the data which were collected from ICU only, and the predict tasks include mortality prediction, length of stay prediction. We ignore the admission prediction, since it requires additional text feature for training with large language model, which is out of scope for this work.

\begin{table}
 \caption{Data extraction and preprocessing configuration}
  \centering
  \begin{tabular}{lll}
    %\toprule
    %\multicolumn{2}{c}{Part}                   \\
    \cmidrule(r){1-2}
     \textbf{Data Extraction}     & \textbf{Data Preprocessing} \\
    \midrule
    1.	MIMIC-IV 2.2  & 1.	Group Diagnoses codes for with ICD-10     \\
    2.	ICU & 2.	Produce data summary for features     \\
    3.	Mortality first 48h, LOS >=3       & 3.	Feature selection for Labs/Vitals  \\
    4.	Admitted for Chronic Kidney Disease & 4.	Outlier removal with threshold 0.98 \\
    5.	Diagnosis and Labs/Vitals & 5.	2-hour time series resolution \\
    & 6.	Forward fill and mean imputation for labs/vitals \\
    & 7.	First 48 hours of admission data \\
    \bottomrule
  \end{tabular}
  \label{tab:table1}
\end{table}
Table~\ref{tab:table1} illustrates the detailed data extraction and preprocessing steps for the pipeline, which is proposed, and experimented by \citet{gupta2022extensive}, and fully provided by the pipeline. During data extraction steps, we used only the data collected from ICU, and admitted for Chronic Kidney Disease only. We also select the features from Diagnosis and Labs/Vitals tables, which contains 407 labs/vitals features, 1034 diagnosis features for mortality prediction task, and 408 labs/vitals features, 1117 diagnosis features for length of stay prediction task. In the data preprocessing step, the pipeline supports clinical grouping for diagnosis and medications features, which helps reduce data dimensionality while preserving important clinical information. The International Classification of Diseases (ICD) is a medical classification list by World Health Organization, containing code for diseases, signs, and symptom, etc. The MIMIC-IV dataset supports both ICD-9 and ICD-10, in which ICD-10 is its 10th revision, which is the latest.  Prior to any grouping of diagnosis data, the ICD-9 codes are converted to ICD-10, so that all codes can be grouped by the same standard.  After grouping, the pipeline provides summaries of the features extracted. These summaries are provided in CSV files, which contain information about the mean frequency and the percentage of missing values for different lab and/or medication codes. The number of admissions is utilized for mortality and length of stay training, are 7910 and 12604 respectively.
\subsection{Prediction Algorithms}
\begin{table}
 \caption{Hyperparameters/Parameters of predictive models}
  \centering
  \begin{tabular}{ |p{3cm}|p{12cm}| }
    %\toprule
    %\multicolumn{2}{c}{Part}                   \\
    %\toprule
    \hline
     \textbf{Model}     & \textbf{Hyper-parameters/Parameters} \\
    %\cmidrule(r){1-2}
    \hline
    XGBoost  & objective=binary:logistic      \\
    %\cmidrule(r){1-2}
    \hline
    LSTM & Input size = 152, hidden size = 256, layers = 2, d/o = 0.8, learning rate = 0.001, batch size = 128, epoch = 20     \\
    %\cmidrule(r){1-2}
    \hline
    TCN       & Layers=4, filters = 128, kernel size = 10, d/o = 0.8, learning rate = 0.001, batch size = 128, epoch = 20 \\
    %\bottomrule
    \hline
  \end{tabular}
  \label{tab:table2}
\end{table}
We compare several algorithms including a traditional machine learning model, i.e xgboost and two complex deep learning models which are convolutional neural network as well as recurrent neural network. Here is the list of models:
\begin{itemize}
\item XGBoost (Extreme Gradient Boosting): Use gradient boosting to combine predictions of multiple weak models, which operates on decision trees, constructing a graph of if conditions to make prediction. \citet{10.1145/2939672.2939785}
\item LSTM (Long Short-Term Memory):  Has two mechanisms, i.e hidden, and internal cell. The hidden state corresponds to the Short-Term Memory (STM) component, and the cell state corresponds to the Long-Term Memory (LTM) \citet{10.1162/neco.1997.9.8.1735}.
\item TCN (Temporal Convolution Network): The convolutions are causal, and model can take a sequence of any length and map it to an output sequence of the same length with an RNN \citet{8099596}.
\end{itemize}
Table ~\ref{tab:table2} summarizes the hyperparameters of the two deep learning models for our experiments. 

\subsection{Result}
We use 5-folded cross validation to verify the model performance, in which the dataset is spit 80\% for training, and 20\% for testing. For deep learning model training, we split the training data further by taking 10\% for validation at the end of each training epoch.

As can be seen from Table ~\ref{tab:table3}, XGBoost achieve the best scores in both mortality and length of stay prediction tasks. On the other hand, LTSM and TCN performance are quite similar in which LSTM outperforms TCN at mortality task but falls behind TCN at length of stay task.
\begin{table}
 \caption{Performance of different models with MIMIC-IV on 2 tasks: in-icu mortality, length of stay.}
  \centering
    %\begin{tabular}{ |c|c|c|c|c| }
    \begin{tabular}{ |p{2.5cm}|p{2.5cm}|p{2.5cm}|p{2.5cm}|p{2.5cm}| }
    %\toprule
    %\multicolumn{2}{c}{Part}                   \\
    %\toprule
    \hline
     \textbf{Task} & \multicolumn{2}{|c|}{\textbf{Mortality}} & \multicolumn{2}{|c|}{\textbf{Length of Stay (LOS>3)}} \\
    %\cmidrule(r){1-2}
    \hline
    \textbf{Metrics} & \textbf{AUC-ROC}  & \textbf{PR-ROC} & \textbf{AUC-ROC}  & \textbf{PR-ROC}     \\
    %\cmidrule(r){1-2}
    \hline
    XGBoost & \textbf{0.87}  & \textbf{0.52} & \textbf{0.76}  & \textbf{0.73}     \\
    %\cmidrule(r){1-2}
    \hline
    LSTM & 0.83  & 0.45 & 0.72  & 0.69     \\
    \hline
    TCN & 0.82  & 0.43 & 0.73  & 0.73     \\
    %\bottomrule
    \hline
  \end{tabular}
  \label{tab:table3}
\end{table}
\section{Conclusion}
This work is aimed to provide a standardized benchmark for MIMIC-IV, one of the most popular EHRs datasets. Although much research are focusing on applying complex deep learning algorithms to MIMIC dataset, traditional machine learning model like XGBoost still able to outperform, and achieve good results in mortality and length of stay prediction tasks. In the future, we will keep updating this work by adding more algorithms, especially state of the art ones into the benchmark, and also extend it with more prediction tasks such as phenotyping and admission. 
%\section*{Acknowledgments}
%This was was supported in part by......

%Bibliography
%\bibliographystyle{unsrt}  
\bibliography{references}

\begin{thebibliography}{23}
\providecommand{\natexlab}[1]{#1}
\providecommand{\url}[1]{\texttt{#1}}
\expandafter\ifx\csname urlstyle\endcsname\relax
  \providecommand{\doi}[1]{doi: #1}\else
  \providecommand{\doi}{doi: \begingroup \urlstyle{rm}\Url}\fi

\bibitem[Shukla and Marlin(2022)]{shukla2022}
Satya~Narayan Shukla and Benjamin~M. Marlin.
\newblock Heteroscedastic temporal variational autoencoder for irregularly sampled time series.
\newblock In \emph{International Conference on Learning Representations}, 2022.
\newblock URL \url{https://openreview.net/forum?id=Az7opqbQE-3}.

\bibitem[Tipirneni et~al.(2022)Tipirneni, Sindhu, Reddy, and K.]{tipir2022}
Tipirneni, Sindhu, Reddy, and Chandan K.
\newblock Self-supervised transformer for sparse and irregularly sampled multivariate clinical time-series.
\newblock \emph{ACM Trans. Knowl. Discov. Data}, 16\penalty0 (6), jul 2022.
\newblock ISSN 1556-4681.
\newblock \doi{10.1145/3516367}.
\newblock URL \url{https://doi.org/10.1145/3516367}.

\bibitem[Shukla and Marlin(2021)]{shukla2021multitime}
Satya~Narayan Shukla and Benjamin Marlin.
\newblock Multi-time attention networks for irregularly sampled time series.
\newblock In \emph{International Conference on Learning Representations}, 2021.
\newblock URL \url{https://openreview.net/forum?id=4c0J6lwQ4_}.

\bibitem[Wang et~al.(2020)Wang, McDermott, Chauhan, Ghassemi, Hughes, and Naumann]{10.1145/3368555.3384469}
Shirly Wang, Matthew B.~A. McDermott, Geeticka Chauhan, Marzyeh Ghassemi, Michael~C. Hughes, and Tristan Naumann.
\newblock Mimic-extract: a data extraction, preprocessing, and representation pipeline for mimic-iii.
\newblock In \emph{Proceedings of the ACM Conference on Health, Inference, and Learning}, CHIL '20, page 222–235, New York, NY, USA, 2020. Association for Computing Machinery.
\newblock ISBN 9781450370462.
\newblock \doi{10.1145/3368555.3384469}.
\newblock URL \url{https://doi.org/10.1145/3368555.3384469}.

\bibitem[Tang et~al.(2020)Tang, Davarmanesh, Song, Koutra, Sjoding, and Wiens]{10.1093/jamia/ocaa139}
Shengpu Tang, Parmida Davarmanesh, Yanmeng Song, Danai Koutra, Michael~W Sjoding, and Jenna Wiens.
\newblock {Democratizing EHR analyses with FIDDLE: a flexible data-driven preprocessing pipeline for structured clinical data}.
\newblock \emph{Journal of the American Medical Informatics Association}, 27\penalty0 (12):\penalty0 1921--1934, 10 2020.
\newblock ISSN 1527-974X.
\newblock \doi{10.1093/jamia/ocaa139}.
\newblock URL \url{https://doi.org/10.1093/jamia/ocaa139}.

\bibitem[Pollard et~al.(2018)Pollard, Johnson, Raffa, Celi, Mark, and Badawi]{Pollard2018TheEC}
Tom~J. Pollard, Alistair E.~W. Johnson, Jesse~Daniel Raffa, Leo~Anthony Celi, Roger~G. Mark, and Omar Badawi.
\newblock The eicu collaborative research database, a freely available multi-center database for critical care research.
\newblock \emph{Scientific Data}, 5, 2018.
\newblock URL \url{https://api.semanticscholar.org/CorpusID:52182706}.

\bibitem[Jarrett et~al.(2021)Jarrett, Yoon, Bica, Qian, Ercole, and van~der Schaar]{jarrett2021clairvoyance}
Daniel Jarrett, Jinsung Yoon, Ioana Bica, Zhaozhi Qian, Ari Ercole, and Mihaela van~der Schaar.
\newblock Clairvoyance: A pipeline toolkit for medical time series.
\newblock In \emph{International Conference on Learning Representations}, 2021.
\newblock URL \url{https://openreview.net/forum?id=xnC8YwKUE3k}.

\bibitem[Mandyam et~al.(2021)Mandyam, Yoo, Soules, Laudanski, and Engelhardt]{10.1145/3459930.3469536}
Aishwarya Mandyam, Elizabeth~C. Yoo, Jeff Soules, Krzysztof Laudanski, and Barbara~E. Engelhardt.
\newblock Cop-e-cat: cleaning and organization pipeline for ehr computational and analytic tasks.
\newblock In \emph{Proceedings of the 12th ACM Conference on Bioinformatics, Computational Biology, and Health Informatics}, BCB '21, New York, NY, USA, 2021. Association for Computing Machinery.
\newblock ISBN 9781450384506.
\newblock \doi{10.1145/3459930.3469536}.
\newblock URL \url{https://doi.org/10.1145/3459930.3469536}.

\bibitem[Gupta et~al.(2022)Gupta, Gallamoza, Cutrona, Dhakal, Poulain, and Beheshti]{gupta2022extensive}
Mehak Gupta, Brennan Gallamoza, Nicolas Cutrona, Pranjal Dhakal, Raphael Poulain, and Rahmatollah Beheshti.
\newblock {An Extensive Data Processing Pipeline for MIMIC-IV}.
\newblock In \emph{Proceedings of the 2nd Machine Learning for Health symposium}, volume 193 of \emph{Proceedings of Machine Learning Research}, pages 311--325. PMLR, 28 Nov 2022.
\newblock URL \url{https://proceedings.mlr.press/v193/gupta22a.html}.

\bibitem[Liao and Voldman(2023)]{LIAO2023104356}
Wei Liao and Joel Voldman.
\newblock A multidatabase extraction pipeline (metre) for facile cross validation in critical care research.
\newblock \emph{Journal of Biomedical Informatics}, 141:\penalty0 104356, 2023.
\newblock ISSN 1532-0464.
\newblock \doi{https://doi.org/10.1016/j.jbi.2023.104356}.
\newblock URL \url{https://www.sciencedirect.com/science/article/pii/S1532046423000771}.

\bibitem[Lea et~al.(2017)Lea, Flynn, Vidal, Reiter, and Hager]{8099596}
C.~Lea, M.~D. Flynn, R.~Vidal, A.~Reiter, and G.~D. Hager.
\newblock Temporal convolutional networks for action segmentation and detection.
\newblock In \emph{2017 IEEE Conference on Computer Vision and Pattern Recognition (CVPR)}, pages 1003--1012, Los Alamitos, CA, USA, jul 2017. IEEE Computer Society.
\newblock \doi{10.1109/CVPR.2017.113}.
\newblock URL \url{https://doi.ieeecomputersociety.org/10.1109/CVPR.2017.113}.

\bibitem[Bednarski et~al.(2022)Bednarski, Singh, Zhang, Jones, Naeim, and Ramezani]{tcn_mimic3}
Bryan~P. Bednarski, Akash~Deep Singh, Wenhao Zhang, William~M. Jones, Arash Naeim, and Ramin Ramezani.
\newblock Temporal convolutional networks and data rebalancing for clinical length of stay and mortality prediction.
\newblock In \emph{Scientific Report 12}. Nature Publishing Group UK, 2022.
\newblock \doi{https://doi.org/10.1038/s41598-022-25472-z}.

\bibitem[Baytas et~al.(2017)Baytas, Xiao, Zhang, Wang, Jain, and Zhou]{10.1145/3097983.3097997}
Inci~M. Baytas, Cao Xiao, Xi~Zhang, Fei Wang, Anil~K. Jain, and Jiayu Zhou.
\newblock Patient subtyping via time-aware lstm networks.
\newblock In \emph{Proceedings of the 23rd ACM SIGKDD International Conference on Knowledge Discovery and Data Mining}, KDD '17, page 65–74, New York, NY, USA, 2017. Association for Computing Machinery.
\newblock ISBN 9781450348874.
\newblock \doi{10.1145/3097983.3097997}.
\newblock URL \url{https://doi.org/10.1145/3097983.3097997}.

\bibitem[Rubanova et~al.(2019)Rubanova, Chen, and Duvenaud]{NEURIPS2019_42a6845a}
Yulia Rubanova, Ricky T.~Q. Chen, and David~K Duvenaud.
\newblock Latent ordinary differential equations for irregularly-sampled time series.
\newblock In H.~Wallach, H.~Larochelle, A.~Beygelzimer, F.~d\textquotesingle Alch\'{e}-Buc, E.~Fox, and R.~Garnett, editors, \emph{Advances in Neural Information Processing Systems}, volume~32. Curran Associates, Inc., 2019.
\newblock URL \url{https://proceedings.neurips.cc/paper_files/paper/2019/file/42a6845a557bef704ad8ac9cb4461d43-Paper.pdf}.

\bibitem[Che et~al.(2018)Che, Purushotham, Cho, Sontag, and Liu]{rnn_scir8}
Zhengping Che, Sanjay Purushotham, Kyunghyun Cho, David Sontag, and Yan Liu.
\newblock Recurrent neural networks for multivariate time series with missing values.
\newblock In \emph{Scientific Report 8}. Scientific Report, 2018.
\newblock \doi{https://doi.org/10.1038/s41598-018-24271-9}.

\bibitem[Tan et~al.(2020)Tan, Ye, Yang, Liu, Ma, Yip, Wong, and Yuen]{Tan_Ye_Yang_Liu_Ma_Yip_Wong_Yuen_2020}
Qingxiong Tan, Mang Ye, Baoyao Yang, Siqi Liu, Andy~Jinhua Ma, Terry Cheuk-Fung Yip, Grace Lai-Hung Wong, and PongChi Yuen.
\newblock Data-gru: Dual-attention time-aware gated recurrent unit for irregular multivariate time series.
\newblock \emph{Proceedings of the AAAI Conference on Artificial Intelligence}, 34\penalty0 (01):\penalty0 930--937, Apr. 2020.
\newblock \doi{10.1609/aaai.v34i01.5440}.
\newblock URL \url{https://ojs.aaai.org/index.php/AAAI/article/view/5440}.

\bibitem[Purushotham et~al.(2018)Purushotham, Meng, Che, and Liu]{sanjay2018}
Sanjay Purushotham, Chuizheng Meng, Zhongping Che, and Yan Liu.
\newblock Benchmark of deep learning models on large healthcare mimic datasets.
\newblock In \emph{Journal of Biomedical Informatics}, pages 112--134, 2018.

\bibitem[Harutyunyan et~al.(2019)Harutyunyan, Khachatrian, Kale, Steeg, and Galstyan.]{hrayr2019}
Hrayr Harutyunyan, Hrant Khachatrian, David~C. Kale, Greg~Ver Steeg, and Aram Galstyan.
\newblock Multitask learning and benchmarking with clinical time series data.
\newblock In \emph{Scientific Data 6}. Scientific Data, 2019.
\newblock \doi{https://doi.org/10.1038/s41597-019-0103-9}.

\bibitem[Hsu et~al.(2020)Hsu, Karnwal, Mullainathan, Obermeyer, and Tan]{hsu-etal-2020-characterizing}
Chao-Chun Hsu, Shantanu Karnwal, Sendhil Mullainathan, Ziad Obermeyer, and Chenhao Tan.
\newblock Characterizing the value of information in medical notes.
\newblock In Trevor Cohn, Yulan He, and Yang Liu, editors, \emph{Findings of the Association for Computational Linguistics: EMNLP 2020}, pages 2062--2072, Online, November 2020. Association for Computational Linguistics.
\newblock \doi{10.18653/v1/2020.findings-emnlp.187}.
\newblock URL \url{https://aclanthology.org/2020.findings-emnlp.187}.

\bibitem[Johnson et~al.(2023)Johnson, BulgarelliLucas, Pollard, Horng, Celi, and Mark.]{mimic422}
Alistair Johnson, Lucas BulgarelliLucas, Tom Pollard, Steven Horng, Leo~Anthony Celi, and Roger Mark.
\newblock Mimic-iv (version 2.2)., 2023.
\newblock URL \url{https://doi.org/10.13026/6mm1-ek67}.

\bibitem[Nguyen et~al.(2023)Nguyen, Schlegel, Kashyap, Winkler, Huang, Liu, and Lin]{nguyen2023mimicivicd}
Thanh-Tung Nguyen, Viktor Schlegel, Abhinav Kashyap, Stefan Winkler, Shao-Syuan Huang, Jie-Jyun Liu, and Chih-Jen Lin.
\newblock Mimic-iv-icd: A new benchmark for extreme multilabel classification, 2023.

\bibitem[Chen and Guestrin(2016)]{10.1145/2939672.2939785}
Tianqi Chen and Carlos Guestrin.
\newblock Xgboost: A scalable tree boosting system.
\newblock In \emph{Proceedings of the 22nd ACM SIGKDD International Conference on Knowledge Discovery and Data Mining}, KDD '16, page 785–794, New York, NY, USA, 2016. Association for Computing Machinery.
\newblock ISBN 9781450342322.
\newblock \doi{10.1145/2939672.2939785}.
\newblock URL \url{https://doi.org/10.1145/2939672.2939785}.

\bibitem[Hochreiter and Schmidhuber(1997)]{10.1162/neco.1997.9.8.1735}
Sepp Hochreiter and Jürgen Schmidhuber.
\newblock {Long Short-Term Memory}.
\newblock \emph{Neural Computation}, 9\penalty0 (8):\penalty0 1735--1780, 11 1997.
\newblock ISSN 0899-7667.
\newblock \doi{10.1162/neco.1997.9.8.1735}.
\newblock URL \url{https://doi.org/10.1162/neco.1997.9.8.1735}.

\end{thebibliography}

\end{document}